\documentclass[sigconf,nonacm]{acmart}
\usepackage{subcaption}
\usepackage{multirow}
\usepackage{tabularx}

\AtBeginDocument{%
  \providecommand\BibTeX{{%
    \normalfont B\kern-0.5em{\scshape i\kern-0.25em b}\kern-0.8em\TeX}}}

\begin{document}

\title{Unsupervised Video Summarization via Multi-source Features}


\author{Hussain Kanafani$^2$, Junaid Ahmed Ghauri$^1$, Sherzod Hakimov$^1$, Ralph Ewerth$^{1,2}$}
\affiliation{%
  \institution{$^1$TIB -- Leibniz Information Centre for Science and Technology \\ $^2$L3S Research Center, Leibniz Univerity Hannover}
  \city{Hannover}
  \country{Germany}}
\email{hussainkanafani@gmail.com, {junaid.ghauri, sherzod.hakimov, ralph.ewerth}@tib.eu}

\renewcommand{\shortauthors}{Kanafani et al.}


\begin{abstract}
Video summarization aims at generating a compact yet representative visual summary that conveys the essence of the original video. The advantage of unsupervised approaches is that they do not require human annotations to learn the summarization capability and generalize to a wider range of domains. Previous work relies on the same type of deep features, typically based on a model pre-trained on ImageNet data. Therefore, we propose the incorporation of multiple feature sources with chunk and stride fusion to provide more information about the visual content. For a comprehensive evaluation on the two benchmarks \textit{TVSum} and \textit{SumMe}, we compare our method with four state-of-the-art approaches. Two of these approaches were implemented by ourselves to reproduce the reported results. Our evaluation shows that we obtain state-of-the-art results on both datasets, while also highlighting the shortcomings of previous work with regard to the evaluation methodology. Finally, we perform error analysis on videos for the two benchmark datasets to summarize and spot the factors that lead to misclassifications. 


\end{abstract}

\keywords{unsupervised video summarization, multi-source combination, multi-source fusion, deep learning, video analysis}

\maketitle

\section{Introduction}


Driven by the rapid growth of visual content in recent years, videos have become the dominant form of information exchange on the Web. According to Cisco Visual Networking Index~\citep{Globa4430341:online}, the video content grows annually with a rate of 33\%, and it will be responsible for 80\% of the global Internet traffic by 2022. However, it is time-consuming to browse long videos, and it is beneficial and preferable to watch a short and concise summary that conveys the main content of the original video. Therefore, automatic video summarization methods are required to view, search, and retrieve video content efficiently. The development of such models in a supervised fashion requires ground-truth summaries for training. 
However, the acquisition of a large number of ground-truth summaries is difficult, time-consuming, and expensive. Furthermore, the training data might introduce a domain or data bias. For these reasons, researchers focused on unsupervised methods that do not require human supervision and yet being able to generalize on a wider range of domains.
Many approaches have tackled the task of unsupervised video summarization using different methods. Earlier approaches developed static video summaries by applying clustering algorithms to long videos~\citep{de2011vsumm, gygli2014creating}. Deep learning approaches used different Generative Adversarial Networks (GAN) variations along with attention mechanisms~\citep{Apostolidis.2019,Apostolidis.2020,jung2019discriminative,mahasseni2017unsupervised, yuan2019cycle,rochan2019video}. Other methods are trained in a reinforcement learning-based framework with reward functions~\citep{zhao2018hsa, yaliniz2019unsupervised}.

In this paper, we propose a deep learning model for unsupervised video summarization called Multi-Source Chunk and Stride Fusion (MCSF), which investigates the impact of multiple visual representations extracted about visual objects and scene (i.e., places) content. It also uses two temporal constellations of the video features which give the model different perspectives of the video, similar to Jung et al~\citep{jung2019discriminative}. Consequently, three fusion strategies are suggested and evaluated. Comprehensive experiments are conducted to compare our approach with three state-of-the-art unsupervised~\citep{Apostolidis.2019,Apostolidis.2020,jung2019discriminative} methods as well as a reinforcement learning method~\citep{yaliniz2019unsupervised} in a fair manner.
We have discovered issues in the evaluation methodology used by these methods with regard to k-fold cross-validation. Some videos were excluded from the test data splits, whereas other videos were repeated multiple times in the test splits to perform k-fold cross-validation. We provide a new evaluation scheme that solves these problems in the datasets and allows for a fair comparison on the two benchmark datasets: TVSum~\citep{song2015tvsum} and SumMe~\citep{gygli2014creating}. 
The proposed solution yields better performance than previous state-of-the-art methods on both datasets. We share the source code for the evaluation scheme, the proposed model architecture, the re-implemented methods, and the newly generated splits of both datasets with the research community\footnote{ \url{https://github.com/TIBHannover/UnsupervisedVideoSummarization}}.


The remainder of the paper is structured as follows. The previous work in the domain of unsupervised video summarization is discussed in Section~\ref{sec:related_work}. In Section~\ref{sec:methodology}, we describe the unsupervised deep learning architecture that harnesses multi-source feature embeddings along with fusion techniques. Experimental results and the comparison with four state-of-the-art methods are presented in Section~\ref{sec:experiments}, while Section~\ref{sec:conclusion} concludes the paper.

\begin{figure*}
  \centering
  \includegraphics[width=\textwidth]{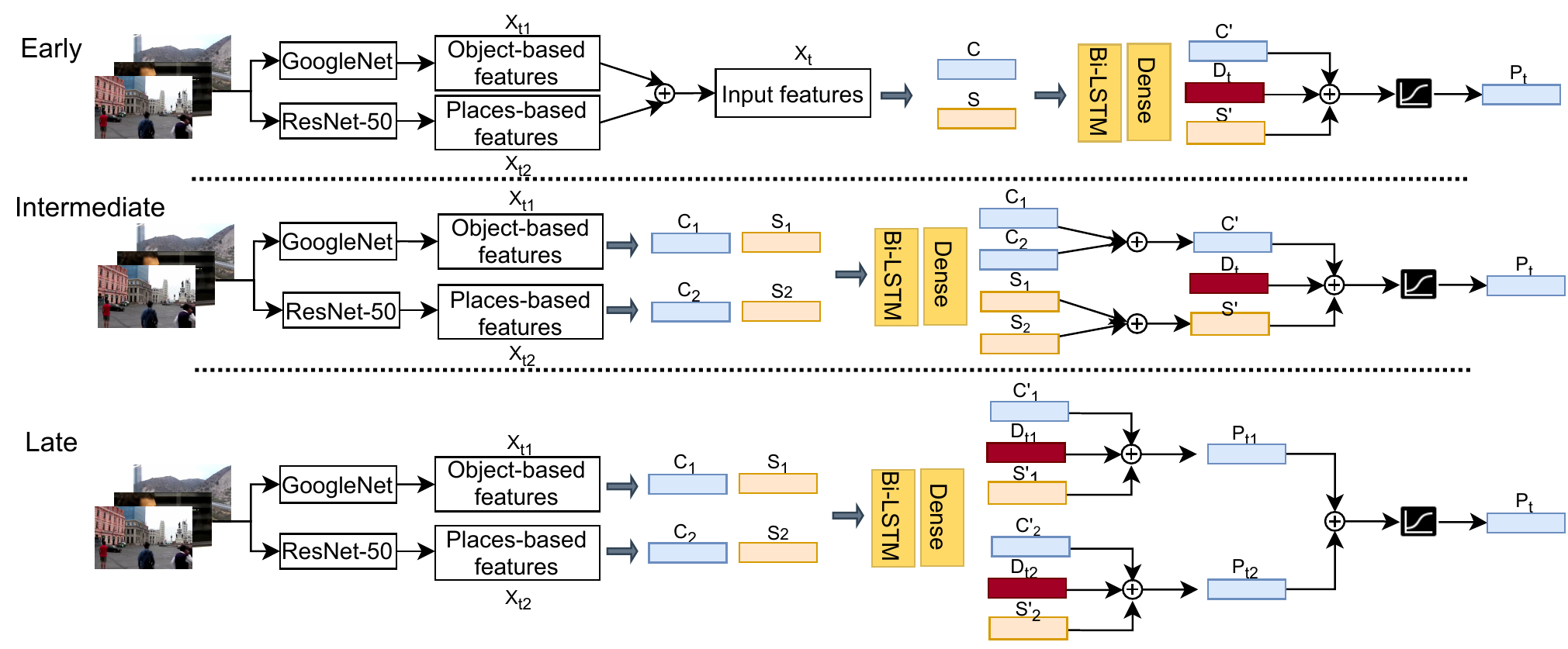}
   \caption{An overview of the Multi-Source Chunk and Stride Fusion (MCSF) architecture with different fusion techniques. $X_{t1}$ and $X_{t2}$ are the extracted features by GoogleNet and ResNet-50 respectively, $C_1$ and $C_2$ are the chunks, $S_1$ and $S_2$ are strides computed similar to ~\citep{jung2019discriminative} and $C^{'}_{1}$, $S^{'}_{1}$, $C^{'}_{2}$, and $S^{'}_{2}$ are the chunks and strides learned by the summarizer. $D_t$ is the difference attention, which is computed for each type of features separately and then summed.}
    \label{fig:methodology}
\end{figure*}

\section{Related Work}\label{sec:related_work}

In recent years, numerous works have approached video summarization from both supervised and unsupervised learning perspectives. Unsupervised methods learn the video summarization capability without the need for ground-truth summaries of videos and aim to generalize on different domains.
The \textit{Video SUMMarization(VSUMM)} approach~\citep{de2011vsumm} is one of the earliest methods for static video summarization, it extracts color features from the video and then performs k-medoid clustering to acquire keyframes. \citet{gygli2014creating} proposed a segmentation-based method that attempts to choose a subset of segments that maximizes the sum of the interestingness of segments. The interestingness score for each segment is computed using a combination of low-level features and high-level information.

Most deep learning methods are trained in a generative adversarial manner to overcome the absence of ground-truth video summaries. One group of methods uses generative adversarial networks (GANs) along with variational autoencoder and Long Short-term Memory (VAE-LSTM)~\citep{mahasseni2017unsupervised,jung2019discriminative, yuan2019cycle, Apostolidis.2019}. Among these methods \textit{SUM-GAN}~\citep{mahasseni2017unsupervised} represents the base that other state-of-the-art adversarial models build upon. \textit{Chunk and Stride Network (CSNet)} ~\citep{jung2019discriminative} uses \textit{SUM-GAN} as a baseline and tackles the problem of gradient decay when dealing with long videos as well as the ineffective feature learning due to flat distribution of frames importance scores. In CSNet, the input features are divided differently into two smaller sequences forming local chunks and global stride (view) of input features. Besides, an attention module is utilized to compute the differences of features of adjacent frames and use them as an indicator of the significance of a specific frame. Another group of methods replace VAE with a deterministic autoencoder (AE) and enhance it with an attention mechanism ~\citep{rochan2019video,Apostolidis.2020}. Another approach formulated it as a decision process where the video summary algorithm is trained using reinforcement learning to find a visual summary that satisfies certain conditions such as uniformity and representativenss. However, only few publications adopt this approach~\citep{zhou2018deep,yaliniz2019unsupervised}. However, the generalizability of the aforementioned methods is not tested, since the reported experiments of most of them were conducted using the canonical settings, i.e., the data come from the same dataset and are split into 80\% for training and 20\% for testing. 

\section{Unsupervised Video Summarization with Multi-Source Features}\label{sec:methodology}

In this section, we describe the overall architecture of the Multi-Source Chunk and Stride Fusion (MCSF) model. 
The overall model architecture is shown in Figure~\ref{fig:methodology}. The frames in videos are sub-sampled at the rate of 
two frames per second where each frame's features are extracted using the respective visual encoder model. Next, the input features are fed through different layers and finally fused in an early, intermediate, or late stage of the pipeline. As depicted in the proposed architecture, chunks and strides are fed to a Bidirectional Long-short Term Memory (Bi-LSTM) and linear layers and the output is summed up with the difference attention. The end results $P_t$ are probabilities to select frames in the summary. The following fusion techniques~\citep{d2015review} have been used to combine multi-source features.

\noindent\textbf{Early fusion}: The fusion is applied at the feature-level where features coming from different sources ($X_{t1}$, $X_{t2}$) are summed first and then fed to the next layers, which continue to compute $P_t$ based on the fused features.

\noindent\textbf{Intermediate fusion}: The fusion is applied after the different chunks and strides are fed through Bi-LSTM and linear layer. The output strides are chunks from the different streams are summed accordingly with the respective difference attention, and passed through the pipeline to compute $P_t$.
     
\noindent\textbf{Late fusion}: The fusion is applied after both resulting importance scores from each computed. The output importance scores $P_{t1}$ and $P_{t2}$ are summed and passed through a sigmoid layer to produce $P_t$. 

The intuition behind the proposed architecture is that compositional approaches can enhance the performance and enable the model to have a better understanding of the video summarization task. Current state-of-the-art methods use feature representations from a single source, mainly object-based features from pre-trained GoogleNet~\citep{szegedy2015going} on ImageNet data~\citep{DBLP:journals/ijcv/RussakovskyDSKS15}. Contrary to prior work, we exploit features from two 
sources to enhance the representation of visual information in frames. Our approach extends previous work~\cite{jung2019discriminative} 
in order to allow for multi-source input features. These features can be fused in different network layers (early, intermediate, late) to obtain frame-level importance scores for summarization. In addition to the object-based features used in previous work, we incorporate features about (visual) scenery and place content in frames using pre-trained ResNet-50~\citep{he2016deep} on Places365~\citep{zhou2017places} dataset. Such a fusion of multiple features allows the model to recognize changes in scene and objects, which are important indicators for summarization. For instance, different \textit{SumMe} videos about a car crossing a railroad ($video_{6}$) and in a desert $video_{23}$, need to have different representations as their scene information is different, and not just rely on the visual representation about object categories present according to an ImageNet model. Thus, our proposed method uses object-based features in combination with scenery and place-related features to capture a richer representation for summarization of videos in an unsupervised fashion. 


\section{Experimental Setup and Results}\label{sec:experiments}
In this section, we present details about the used benchmark datasets, evaluation metrics, comparison with state of the art, discussion of the results, and qualitative video-wise error analysis.

\subsection{Datasets and Evaluation Metrics}
The following two benchmark datasets were used for all experiments:

\begin{itemize}
	\item \textbf{SumMe}~\citep{gygli2014creating} consists of 25 videos, ranging from one to six minutes, annotated by 15 to 18 users\footnote{\url{https://gyglim.github.io/me/vsum/index.html}}.
	\item \textbf{TVSum}~\citep{song2015tvsum} consists of 50 videos, ranging from two to 10 minutes, annotated by 20 users\footnote{\url{https://github.com/yalesong/tvsum}}.
\end{itemize}

\textbf{Methodological issues in evaluations of previous work}: The evaluation of previous approaches on the two datasets is based on $k$-fold cross-validation where $k=5$. When analyzing the structure of the splits used for evaluation, we found that a non-trivial number of the videos are not part of any test set of the five data splits. More specifically, some videos appear in the test data multiple times, and models are evaluated on them, whereas the model's performance is never tested against some other videos. In total, 28\% and 32\% of the videos in \textit{SumMe} and \textit{TVSum} datasets respectively are not in the test data of the evaluated data splits provided by ~\citep{Apostolidis.2019}. 
Therefore, this work uses new randomly generated \textit{non-overlapping} splits to ensure that each video is contained only once in each split and all videos are included for the evaluation.

\textbf{Evaluation metrics}: This work assesses the performance of compared methods using $F_{1}$ scores on different splits of datasets.

\subsection{Results}\label{subsec:results}
\textbf{Ablation Study:} We evaluated the variations of \textit{MCSF} method and measured the effect of fusing multi-source features.
Similar to previous work, we performed the evaluation on \textit{SumMe} dataset by comparing the model predictions with the ground-truth summaries provided by each user and selecting the maximum value (Max). For \textit{TVSum}, the average value (Avg) between ground-truth summaries and model predictions is used. 
As depicted in Table~\ref{tab:fusion_results}, the incorporation of different types of features enhances the overall $F_{1}$ score. The highest performance on \textit{SumMe} is achieved through intermediate fusion, whereas the late fusion performs the best on \textit{TVSum}. Basically, we experimented with different fusion operations such as summation and averaging. However, reported results in Table~\ref{tab:fusion_results} refer to the summation of different types of features as this operation achieved the highest performance. We also tested the model separately using only the object ($O$) or places and scenery ($P$) features without any fusion and fusing both features yielded better performance.

\begin{table}[!t]
\caption{Ablation study of MCSF model using different types of features, namely: Object-based ($O$) and Places and scenery-based features ($P$). The results are based on two different versions of splits: $F_{1}'$ on the re-evaluated provided by ~\citep{Apostolidis.2019}, ($F_{1}^*$) on \textit{non-overlapping} splits proposed in this paper. }
\begin{center}
\begin{tabular}{ |c|c|c|c|c|}
\hline
Dataset & Fusion & Features & $F_{1}'$ & $F_{1}^*$ \\
\hline
\multirow{5}{3.5em}{SumMe} & - & $O$ & \textbf{48.1} & 41.5 \\ 
\cline{2-5} &  - & $P$  & 46.5 & 40.5 \\
\cline{2-5} &  Early  & $O$ + $P$ & 46.9 & 39.6  \\
\cline{2-5} & Intermediate & $O$ + $P$ & 46.0 & \textbf{43.3} \\
\cline{2-5} & Late & $O$ + $P$ & 47.9 &  40.3   \\
\hline
\hline
\multirow{5}{3.5em}{TVSum} & - & $O$ & 56.4 & 53.4  \\
\cline{2-5} & - & $P$  & 54.2 & 53.6 \\
\cline{2-5} & Early & $O$ + $P$ & 54.9 & 54.3  \\
\cline{2-5} & Intermediate & $O$ + $P$ & 55.7 & 53.8\\
\cline{2-5} & Late & $O$ + $P$ & \textbf{59.1} & \textbf{56.5}\\

\hline
\end{tabular}
\label{tab:fusion_results}
\end{center}
\end{table}

\begin{table}[!t]
\caption{The performance of different unsupervised methods with the re-evaluated ($F_{1}'$) and non-overlapping ($F_{1}^*$) splits compared to the reported results ($F_{1}$). 
}
\begin{center}
\resizebox{\columnwidth}{!}{
\begin{tabular}{|c|c|c|c|c|c|c|c|}
\hline
\multirow{2}{*} {Dataset} & \multirow{2}{*}{Method} & \multicolumn{2}{c|}{$F_1$} & \multicolumn{2}{c|}{$F_{1}'$} &  \multicolumn{2}{c|}{$F_{1}^*$} \\
\cline{3-8} \multicolumn{1}{|c|}{} & \multicolumn{1}{c|}{} & Avg & Max & Avg & Max & Avg & Max \\
\hline
\multirow{5}{3em}{SumMe}& SUM-Ind$_{LU}$ \citep{yaliniz2019unsupervised}  & - & \textbf{51.9} & 22.1  & 46.0 & 18.1 & 42.3   \\
 \cline{2-8}& CSNet \citep{jung2019discriminative} & - & 51.3  & 22.7   & \textbf{48.1} &  18.0 & 41.5    \\ 
 \cline{2-8}& SUM-GAN-AAE \citep{Apostolidis.2020} & - & 48.9  & \textbf{22.8} & 47.1 & 19.0 &  39.8  \\
 \cline{2-8}& SUM-GAN-sl \citep{Apostolidis.2019} & - & 47.3  & 20.4 & 44.6
  & 17.7 & 37.7   \\
\cline{2-8}& MCSF \textbf{(ours)}& - & - & 21 & 46.0 & \textbf{20.1} & \textbf{43.3}   \\  
 
\hline
\hline
\multirow{5}{3em}{TVSum}& SUM-Ind$_{LU}$ \citep{yaliniz2019unsupervised}  & \textbf{61.5} & -  & 58.7  & 80.7 & 55.9 &  77.5 \\

 \cline{2-8}& CSNet \citep{jung2019discriminative} & 58.8 & - & 56.4   & 77.7 & 53.4 &  76.2  \\
 \cline{2-8}& SUM-GAN-AAE \citep{Apostolidis.2020} & 58.3 & - & 57.7   & \textbf{81.6}  & 55.1 &  \textbf{77.8}  \\
 \cline{2-8}& SUM-GAN-sl \citep{Apostolidis.2019} & 58.0 & - & 57.4  & 81.1  & 54.5 &  77.4  \\
\cline{2-8}& MCSF \textbf{(ours)}& - & - & \textbf{59.1} & 81.2 & \textbf{56.5} & 77.6  \\  
\hline
\end{tabular}}
\label{tab:overall_comparison}
\end{center}
\end{table}

\textbf{Overall Comparison}:
The evaluation results for the different state-of-the-art methods are listed in Table~\ref{tab:overall_comparison} and compared with our method. The implementations for SUM-GAN-sl~\citep{Apostolidis.2019} and SUM-GAN-AAE~\citep{Apostolidis.2020} methods are provided. We implemented both CSNet~\citep{jung2019discriminative} and SUM-Ind$_{LU}$ ~\citep{yaliniz2019unsupervised} since the authors did not provide their source code. The compared methods are evaluated on two versions of the five-fold data splits of both benchmark datasets using average $F_{1}$ scores using cross-validation. We computed the $F_{1}$ scores using both Average and Maximum approaches for both datasets. We also included the reported results ($F_{1}$), re-implemented and evaluation results on splits provided by ~\citep{Apostolidis.2019} ($F_{1}'$), and results obtained using the \textit{non-overlapping} splits proposed in this paper ($F_{1}^*$). We have included the best combination of our model that uses both visual features with intermediate and late fusion techniques for \textit{SumMe} and \textit{TVSum} datasets, respectively. It can be seen that the \textit{MCSF} achieved higher performance on the \textit{non-overlapping} splits of both datasets, compared to other methods. Moreover, we analyzed the results obtained and observed a drop in performance for all methods when the non-overlapping splits are used.


\subsection{Error Analysis}
\label{sec:error_analysis}
\begin{figure}[!ht]
       \centering
    \begin{subfigure}[b]{1\textwidth}
     \includegraphics[width=.49\textwidth]{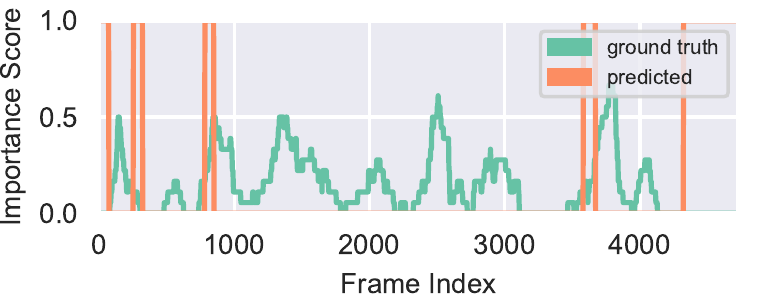}
   \label{fig:csnet_summe_video_2}
   \end{subfigure}
   \begin{subfigure}[b]{0.48\textwidth}
     \includegraphics[width=\textwidth]{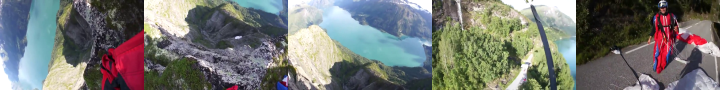}
   \label{fig:csnet_summe_video_2}

   \end{subfigure}
   \begin{subfigure}[b]{.49\textwidth}
     \includegraphics[width=\textwidth]{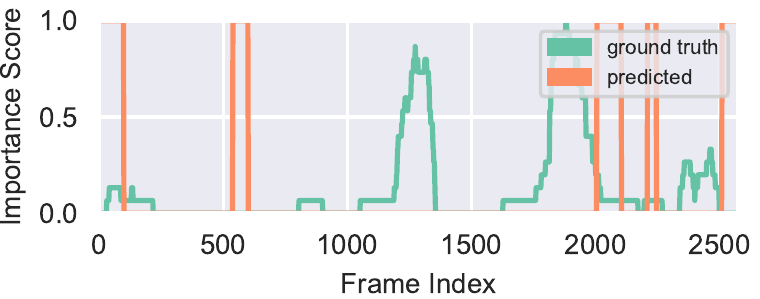}
   \label{fig:csnet_summe_video_24}
   \end{subfigure}
   \begin{subfigure}[b]{0.48\textwidth}
     \includegraphics[width=\textwidth]{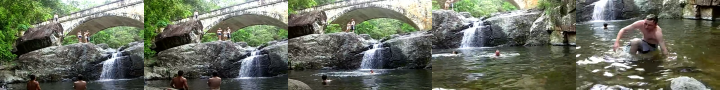}
   \label{fig:csnet_summe_video_24}

   \end{subfigure}
   \caption{
The upper illustrations demonstrate the generated summary of $video_2$(Base Jumping) and $video_{24}$(Paluma Jump) using the MCSF model compared to the mean of the reference summaries. The lower images show the centers of the parts selected by the generated summary. In the upper illustration, the model is confused by the abrupt visual changes at the end of $video_2$. In the lower illustration, the model does not capture the moving object in the background of $video_{24}$. In both videos, the model fails to capture the jumping event.}

 \label{fig:error_analysis}
 \end{figure}

In the following, an error analysis is performed for the following video IDs in \textit{SumMe} dataset: 2, 5, 6, 7, 10, 12, 14, 17, 21, 24. Our goal is to find out what causes the predicted summaries to be not meaningful and fail to convey the essence of the videos. We grouped the issues under five categories as follows. Sample videos with their predictions compared with ground-truth summaries are shown in Figure~\ref{fig:error_analysis}.

\noindent\textbf{Abrupt visual changes}: Video segments with considerable visual changes, including shaking camera, are taken more into account in the end summary, even though these parts may not represent a relevant action related to the story of the video. 
	  
\noindent\textbf{Activities with inconsiderable visual changes}: In contrast to the previous point, video segments with stagnation or trivial changes in visual features are being discarded from the generated summary, although these segments may contain an essential part of the video's story or content. 
	
\noindent\textbf{Long-temporal activities}: Activities that are distributed over multiple shots are hard to capture by all state-of-the-art approaches since the methods can not infer which part of the activity is more representative and what its boundaries are. 
	
\noindent\textbf{Moving objects in the background}: Current models are confused by scenes where the camera is filming a moving object. 
      
\noindent\textbf{Unrecognized certain actions}: Current state-of-the-art models fail to detect essential actions such as jumping, car crash, and landing. These types of actions can be fundamental for the entire video, and discarding them makes the summary incomplete. 

\subsection{Discussion}\label{subsec:discussion}
Overall, results obtained from unsupervised methods on the original 
splits were close to the reported ones.
Yet, the many videos that were from the test splits (and other videos evaluated twice) led to an unfair evaluation. 
The error analysis determined that existing methods have difficulties with videos filmed using moving camera settings. 
These difficulties can be attributed to two main reasons. First, the evaluated methods are basically trained using only object-based features that process only frame-level information.
Second, those methods create a representative summary that has a similar distribution to the original video without considering the relationships between video segments. 
Our approach addressed the first issue and presents a corresponding solution. 

\section{Conclusions}\label{sec:conclusion}
In this paper, we evaluated the state-of-the-art unsupervised video summarization methods and proposed a solution to bridge the existing gaps.
Therefore, we have proposed multi-source features with chunk and stride fusion to provide more information about the visual content. 
For the evaluation task, we re-implemented two methods and reproduced their reported results. Furthermore, all the methods are fairly compared using two different evaluation metrics and different kinds of splits. 
By applying the late fusion variation on \textit{TVSum}, our approach achieved better results than the state-of-the-art methods when using the fair evaluation scheme with re-organized data splits. 
For \textit{SumMe}, there was an improvement on the non-overlapping splits with an intermediate fusion. 
Eventually, observations of the existing methods were made based on the obtained results, and a detailed video-wise qualitative analysis on the causes for the current shortcomings was conducted.
In the future, we will explore the incorporation of features from actions and other modalities to enhance model performance.

\newpage  

\bibliographystyle{ACM-Reference-Format}

\bibliography{references}

\end{document}